\documentclass[10pt,twocolumn,letterpaper]{article}

\usepackage[pagenumbers]{cvpr}      

\usepackage{amsmath,amssymb,booktabs,tabularx,graphicx,xspace,xurl}
\usepackage{caption}

\providecommand{\method}{\textsc{PIVOT}\xspace}

\providecommand{\robot}{Unitree G1}  
\providecommand{\trackerhz}{50}      
\providecommand{\replanhz}{10}       

\definecolor{cvprblue}{rgb}{0.21,0.49,0.74}
\usepackage[pagebackref,breaklinks,colorlinks,allcolors=cvprblue]{hyperref}
\usepackage{caption}

\def\confName{CVPR}
\def\confYear{2027}

\title{PIVOT: Learning Cooperative Humanoid Loco-Manipulation\\ in Confined Spaces}

\author{%
  Amir Hossain Raj \quad
  Dibyendu Das$^{*}$ \quad
  Harsh Rangwala$^{*}$ \quad
  Xuesu Xiao
}

\begin{document}
\twocolumn[{%
\renewcommand\twocolumn[1][]{#1}%
\maketitle

\vspace{-2em}
\begin{center}
  \captionsetup{type=figure}
  \includegraphics[width=\textwidth]{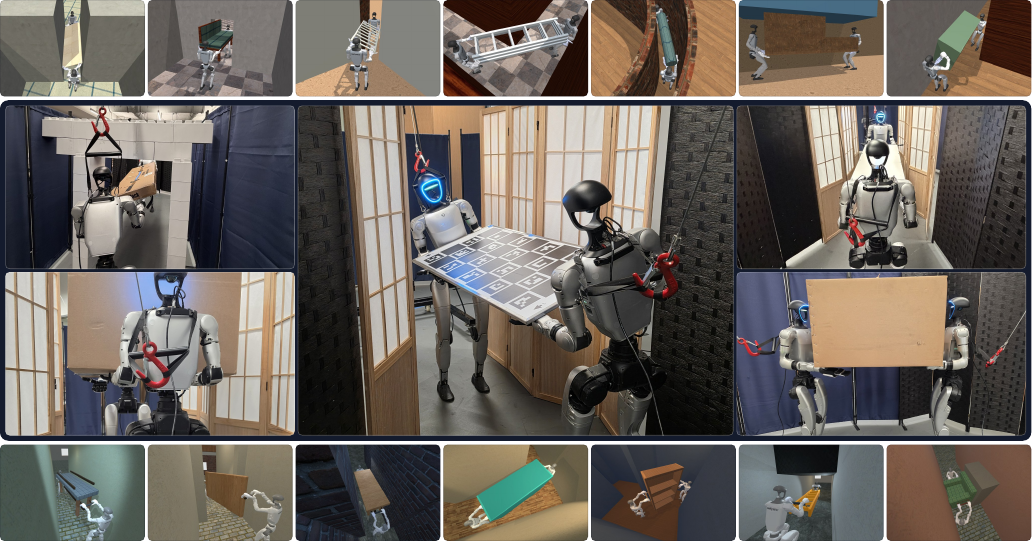}
  \caption{\textbf{Center:} two \robot{} humanoids carrying diverse loads through confined corridors with \method{}. \textbf{Top:} simulated scenes used for training. \textbf{Bottom:} VR scenes used for paired human demonstrations.}
  \label{fig:teaser}
\end{center}%
}]
\begingroup
\renewcommand{\thefootnote}{}
\footnotetext{$^{*}$Equal contribution. All authors are with the Department of Computer Science, George Mason University
        {\tt\footnotesize \{araj20, ddas6, hrangwa, xiao\}@gmu.edu}}
\endgroup
\begin{abstract}
While humanoid robots can walk, carry, and manipulate on their own, many everyday
objects, from ladders to dining tables, can be too large for one
robot to transport alone and require two to work as a team.
In this paper, we develop a learning framework for
vision-based cooperative transport, allowing two humanoid robots to carry long
and large objects through confined passages. Our approach treats the two robots as a single system built around the load they share: their views are fused in a frame attached to the load, and one policy plans its motion together with both carriers. A unified training pipeline
combines simulation, human demonstrations, and real-robot data
to supervise one centralized policy, which is then refined with
reinforcement learning. Deployed on two humanoid robots, the
resulting policy transports diverse objects through corridors
with turns, narrowings, and low ceilings, and generalizes to
layouts and objects not seen in training.
\end{abstract}    
\section{Introduction}
\label{sec:intro}

Humanoid robots have demonstrated increasingly sophisticated
loco-manipulation capabilities from onboard vision, enabled by
advances in whole-body motion tracking~\cite{luo2025sonic,ze2025twist,he2024omnih2o},
visual sim-to-real transfer~\cite{he2026viral,xue2026doorman},
and humanoid foundation models~\cite{bjorck2025gr00t,wei2026psi0}. Yet many
everyday objects in homes, warehouses, and construction sites,
such as doors, bookshelves, and long beams, are too large for
one humanoid to transport alone. Moving these objects requires
multiple robots to work together, extending loco-manipulation
from individual autonomy to physical collaboration.

This challenge becomes particularly demanding in human-centered
environments, where corridors, doorways, sharp corners, and low
ceilings leave little room for maneuvering. Carrying a long
object through such spaces requires more than moving toward a
goal. The robots may need to rotate the load around a corner,
tilt it through a narrow opening, or lower it beneath an
obstacle, while continuously adjusting their own positions
to accommodate the motion. Because both robots hold the same
object, the movement of one directly affects the other.
Perception is equally challenging: the load and the partner
obstruct each robot's view, and critical parts of the surrounding
geometry may be visible to only one of them. Successful
transport therefore requires the robots to perceive their
surroundings and coordinate their movements as a team.

Prior work has made important progress toward cooperative
transport. Physics-based character methods learn multi-humanoid
carrying in simulation from privileged
state~\cite{gao2024coohoi,lionar2026teamhoi}. Real humanoid
systems track commanded object motion with external object
localization~\cite{pandit2026decmht}, or collaborate with
human partners who participate in the
task~\cite{agravante2019human,bethala2025h2compact,du2025cola,zhang2026c2c}.
Earlier two-humanoid systems demonstrated obstacle-aware
transport using onboard cameras through explicit mapping,
planning, and synchronization
pipelines~\cite{rioux2017cooperative,hawley2019control}.
Meanwhile, perceptive controllers enable individual humanoids
to squeeze, crouch, and step through constrained
environments~\cite{xue2026humanoidpf,passage2026,ben2026gallant}.
Together, these advances provide important foundations, but
learning to coordinate two physically connected humanoids
through confined spaces using onboard perception remains
an open challenge.

We introduce \method{}, a learning framework for vision-based
cooperative humanoid transport through confined spaces (Fig.~\ref{fig:teaser}).
Our key intuition is that the robots should reason about
the object they carry and their own movements as one system.
Rather than planning for each robot independently, \method{}
uses the shared load as a common reference for perception
and action. A single policy combines complementary observations
from both robots and predicts how the load and the two carriers
should move together. This formulation enables the robots to
reason about the surrounding space while maintaining coordinated
motion, and provides a common action space for learning from
different sources of experience.

To learn such coordinated behaviors, we collect demonstrations
from pairs of operators who steer virtual humanoids through
confined passages in immersive virtual reality (VR).
These demonstrations capture how people adjust the load and
their movements to negotiate challenging spaces, providing
examples that guide learning in simulation and directly
supervise the vision-based policy. Together with real-robot
teleoperation, they enable us to train a unified policy for
cooperative transport. We evaluate \method{} across seven corridor types in
simulation and validate physical transfer through tight
passages with two \robot{} humanoids. Simulation comparisons
use decentralized, single-view, and planning-based
alternatives, while hardware trials compare against
expert teleoperation. On held-out layouts and loads, we further ablate
shared-load actions, geometric view alignment, load
perception, clearance-guided selection, paired human
demonstrations, and RL post-training.
Our main contributions are:
\begin{itemize}
    \item We present the first learned policy for two humanoids
    to cooperatively transport objects through confined spaces
    using onboard perception alone.

    \item We introduce a shared-load formulation that unifies
    cooperative perception and action for coordinated transport.

    \item We show that paired human demonstrations in VR enable learning of coordinated maneuvers and support transfer to a vision-based policy alongside real-robot teleoperation.
\end{itemize}
\section{Cooperative Transport Around a Shared Load}
\label{sec:method}

\begin{figure*}[t]
  \centering
  \includegraphics[width=\textwidth]{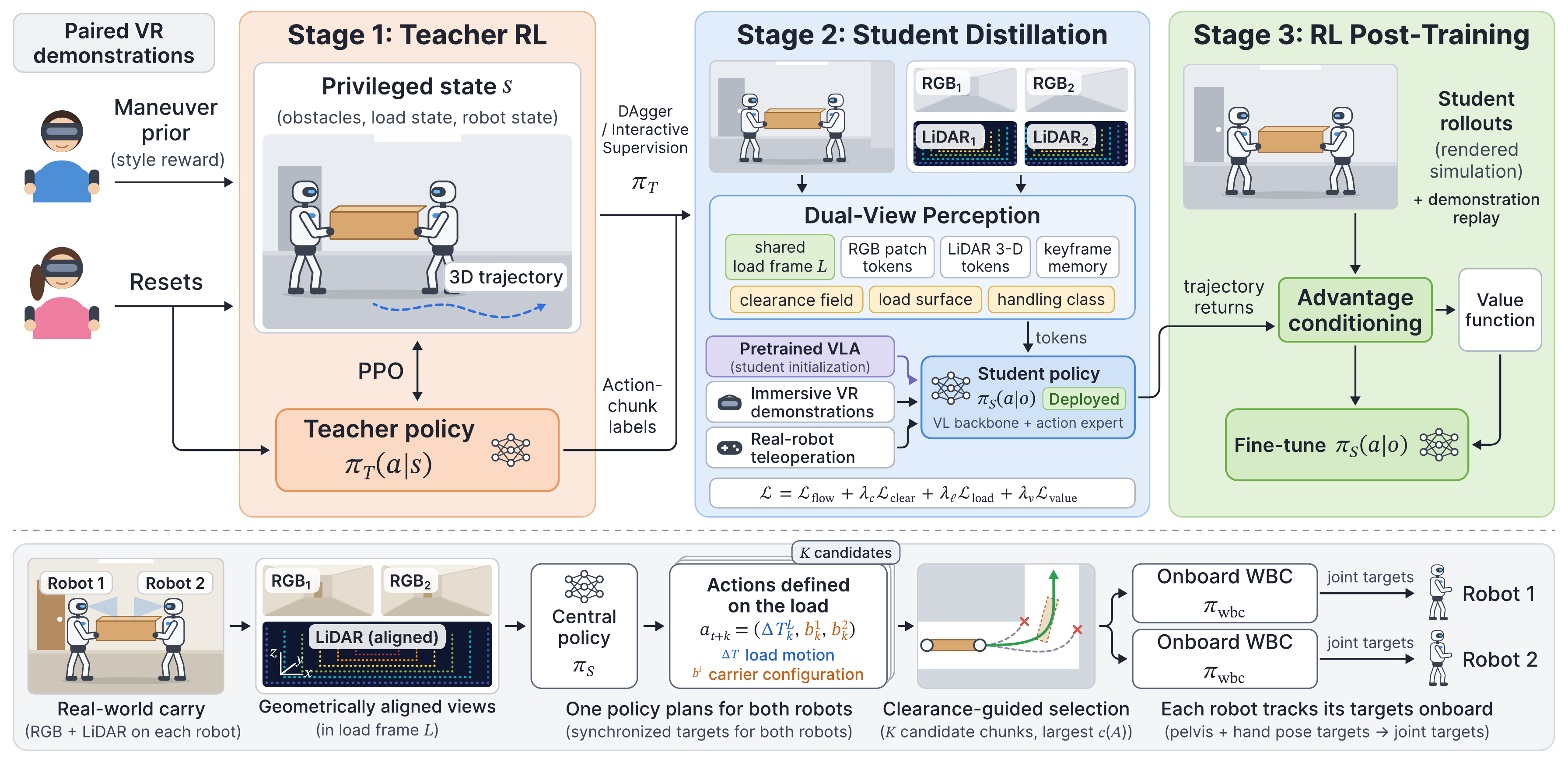}
  \caption{\textbf{\method{} overview.} \textbf{Training:} In Stage~1, paired VR demonstrations guide a privileged RL teacher through a maneuver prior and demonstration-based initial states. In Stage~2, teacher-generated labels, paired VR demonstrations, and real two-robot teleoperation train a vision-based student initialized from a pretrained robot foundation model. Stage~3 refines the student using its own rollouts. \textbf{Deployment:} A single policy fuses both robots' onboard observations, predicts joint motions of the shared load and carriers, and selects a candidate using learned clearance; each robot tracks its targets with an onboard whole-body controller.}
  \label{fig:overview}
\end{figure*}

\textbf{Task.} Two humanoids hold a rigid object and must transport it to a goal region on the far side of a confined passage, without contact between the environment and the load or either robot. At each step, the policy observes a short history of both robots' RGB images, LiDAR scans, and proprioception, together with a goal direction and the grasp configuration, i.e., where the four hands hold the load. Object shape and handling constraints are inferred from the observations. We treat the two robots and the load as one partially observed system, controlled by a single policy that outputs synchronized actions for both robots.

\noindent \textbf{Training.} We learn this policy in three stages (Fig.~\ref{fig:overview}). First, a privileged teacher is trained with RL in simulation, guided by paired VR demonstrations both through a learned maneuver prior and through demonstration-based initial states. Second, the teacher supervises a vision-based student, which also learns directly from VR demonstrations and from teleoperation of the two physical robots. Third, the student is refined with RL on its own simulation rollouts. Human demonstrations thus support teacher exploration and student imitation, while the final policy acts from onboard observations alone.

\noindent \textbf{Deployment.} At each replanning step, both robots provide synchronized onboard observations to the central policy. Their sensor features are aligned relative to the shared load and fused with a short history, allowing information observed by one robot to support the other when the object or partner blocks its view. The policy estimates the load geometry and surrounding clearance, proposes $K$ candidate action chunks, each a short joint motion of the load and both carriers, and selects the candidate with the largest predicted clearance. The chosen chunk provides synchronized targets to a whole-body controller onboard each robot, separating cooperative maneuver planning from lower-level motion tracking.

\subsection{Actions Defined on the Load}
\label{sec:setup}

\begin{figure}[t]
  \centering
  \includegraphics[width=\columnwidth]{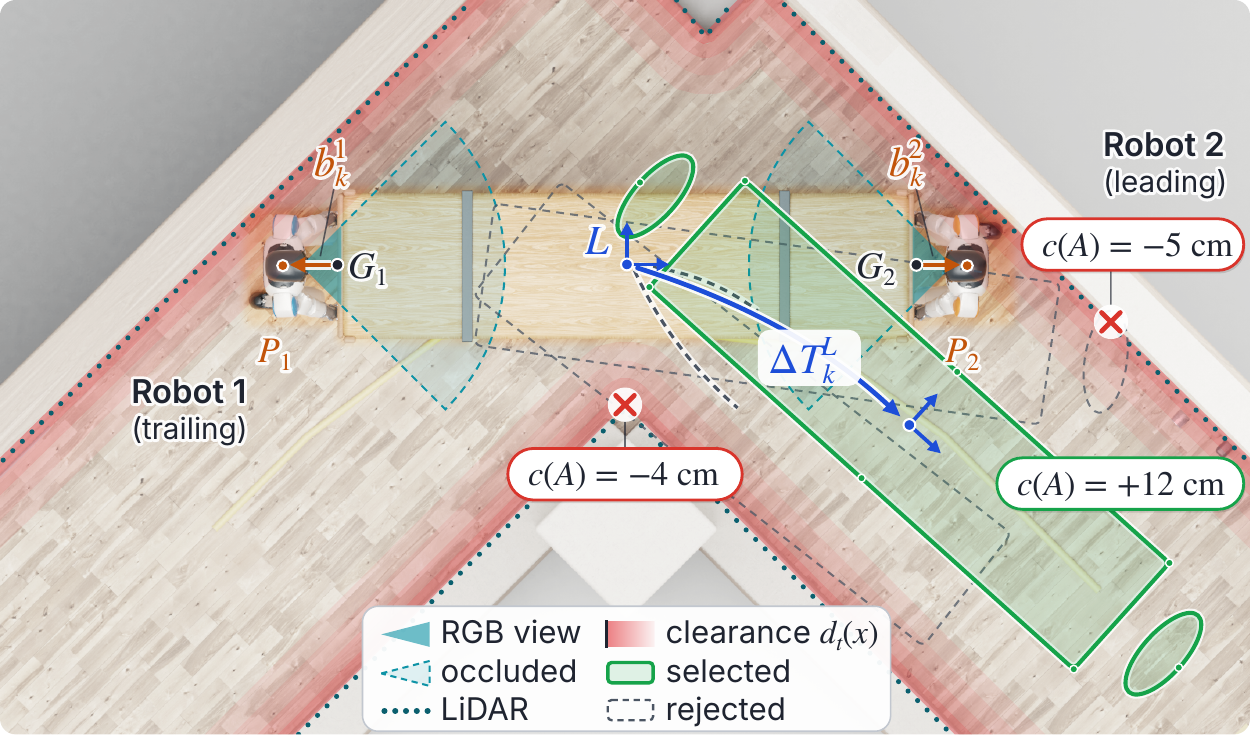}
  \caption{\textbf{Actions and perception in the load frame.} $L$ denotes the shared load frame, $G_1$ and $G_2$ the grasp frames associated with the two hand pairs, and $P_1$ and $P_2$ the robot pelvis frames. The policy predicts load motion (blue) and carrier-relative poses (orange), aligns the two robots' partial observations, and selects the candidate chunk with the largest clearance margin $c(A)$.}
  \label{fig:formulation}
\end{figure}

Each robot $i\in\{1,2\}$ holds the load with both hands. We summarize its two hand contacts by a single grasp frame $G_i$, fixed to the load between that robot's hands. The hand-to-load transforms of all four hands remain fixed during transport. A load frame $L$ is placed midway between $G_1$ and $G_2$ and moves rigidly with the object (Fig.~\ref{fig:formulation}).

The policy predicts the motion of the load and the pose of each carrier relative to its hold. Let $P_i$ be robot $i$'s pelvis frame. For each step $k$ in a chunk of $H$ future actions, the action is
\begin{equation}
  a_{t+k} = \big(\Delta T^L_k,\; b^1_k,\; b^2_k\big),
  \label{eq:action}
\end{equation}
where $\Delta T^L_k\in SE(3)$ is the pose of the load frame at step $t+k$ expressed in the load frame at time $t$, and $b^i_k\in SE(3)$ is the pose of $P_i$ expressed in $G_i$ at step $t+k$. Thus $\Delta T^L_k$ specifies the desired load motion, and $b^i_k$ the corresponding carrier configuration, including orientation, lateral offset, and body height. Both are relative poses, so no world frame is involved. Each of the three poses uses three translation values and a continuous six-dimensional representation of 3-D rotation~\cite{zhou2019continuity}, giving $3\times(3+6)=27$ action values per step. Deriving all four hand targets from a shared load trajectory ensures geometrically consistent commanded grasp poses. A whole-body controller $\pi_{\mathrm{wbc}}$ onboard each robot tracks its pelvis and hand targets, as in recent visual humanoid systems~\cite{he2026viral,xue2026doorman}.

This representation also provides a common label space across our data sources: a privileged RL teacher autonomously controls the simulated robot pair; two human operators each steer one virtual humanoid in VR; and two human operators each teleoperate one physical humanoid. All three sources provide load motion and carrier-relative pose trajectories in the form of Eq.~\eqref{eq:action}.

\subsection{Dual-View Perception and Clearance}
\label{sec:perception}

\textbf{Geometric alignment.} Each robot contributes one RGB camera and one LiDAR, giving two image streams and two LiDAR streams in total. During transport, the fixed hand attachments, robot kinematics, and calibrated sensor extrinsics give the pose of all four sensors in the load frame $L$. The two robots' observations are therefore aligned without external tracking; local odometry is used to update temporal memory and track progress toward the goal.

Each image patch is tagged with the 3-D ray along which it views the scene, namely its camera center and viewing direction expressed in $L$, so the network knows where each patch looks relative to the load. LiDAR measurements are transformed into the same frame and encoded as 3-D spatial tokens. A shared encoder combines the observations of both robots with recent keyframes, retaining evidence from corridor regions that later become occluded.

\textbf{Clearance and load perception.} A clearance head decodes a signed distance field around the team, extending potential fields learned for a single humanoid~\cite{xue2026humanoidpf}:
\begin{equation}
    \big(d_t(x),\,n_t(x)\big)=f_\psi(z_t,x),\qquad x\in\mathbb{R}^3,
  \nonumber
\end{equation}
where $z_t$ is the fused observation representation, $x$ is a query point expressed in $L$, $d_t(x)$ is the signed distance from $x$ to the nearest environment surface (positive in free space), and $n_t(x)$ is the unit direction of increasing signed distance, where the gradient is defined. Simulator geometry provides supervision for both. A separate load head estimates the carried object's surface and its handling class: \emph{free}; \emph{keep level}, which limits how far the load may tilt; or \emph{fragile face}, which requires a larger clearance margin on that face. RGB matters most for the load head: appearance indicates the handling class, which geometry alone does not reveal, and images capture thin parts of the load that range scans can miss. LiDAR provides metric range to the corridor surfaces. These estimates allow the policy to reason about protrusions and motion constraints of loads unseen during training.

\textbf{Clearance-guided selection.} A candidate action chunk $A$ specifies a future load trajectory and the two carriers' target poses, and therefore where every point of the team will be. We score each candidate by its clearance margin $c(A)$: the smallest predicted clearance, less the required safety margin, over all steps of the chunk and over points on the predicted load surface and on a conservative body envelope around each carrier. The policy executes the candidate with the highest $c(A)$ and slows or stops if none is positive (Fig.~\ref{fig:formulation}). Evaluating the entire team, rather than the load alone, accounts for the space needed by both carriers during turns and changes of formation.

\begin{figure}[t]
  \centering
  \includegraphics[width=\columnwidth]{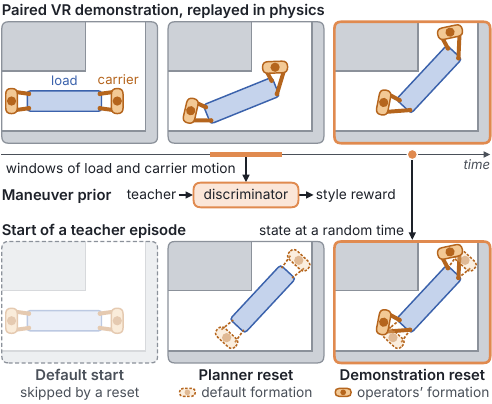}
  \caption{\textbf{Demonstration-guided exploration.} A paired VR demonstration (top) gives the teacher a maneuver prior and reset states that start episodes mid-maneuver (bottom).}
  \label{fig:exploration}
\end{figure}

\begin{figure*}[t]
  \centering
  \includegraphics[width=\textwidth]{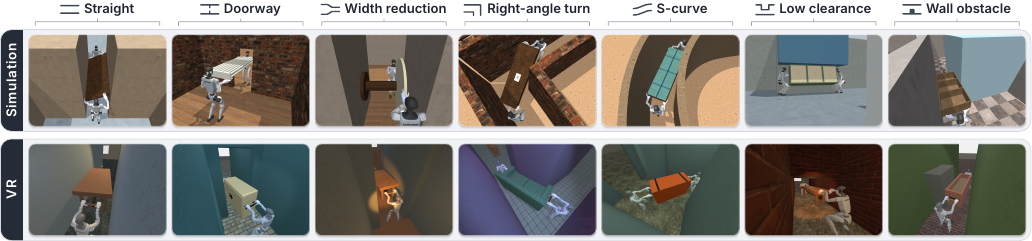}
  \caption{\textbf{Corridors and loads.} Example scenes across environment themes in simulation (top) and VR (bottom).}
  \label{fig:data}
\end{figure*}

\subsection{Learning from Simulation and Human Demonstrations}
\label{sec:learning}

\textbf{Stage 1: privileged teacher RL.} We train a centralized teacher $\pi_T(a\mid s)$ with PPO~\cite{schulman2017ppo} in Isaac Lab~\cite{mittal2025isaaclab}. The teacher controls both robots through the action space of Eq.~\eqref{eq:action} and the whole-body controllers, which are adapted from a pretrained motion tracker~\cite{luo2025sonic} and then frozen. Its privileged state $s$ includes robot and object states, corridor geometry, clearance, and the contact forces and torques at the hands, all read from the simulator. Rewards emphasize progress, completion, safe clearance, low hand forces, and appropriate load handling.

Paired VR demonstrations enter this stage in two complementary ways, both aimed at improving exploration (Fig.~\ref{fig:exploration}). Tight maneuvers are separated from the start state by long stretches without reward: at a corner, the load must be tilted or the formation changed before any progress resumes, and a teacher trained from the start state alone rarely reaches such states.

As a \emph{maneuver prior}, we use an adversarial style reward~\cite{peng2021amp} to make the teacher move more like the demonstrations while it explores. A discriminator is trained to tell short windows of load and carrier motion (the quantities of Eq.~\eqref{eq:action}) in the demonstrations from those produced by the teacher. The teacher receives, in addition to its task reward, a style reward that is higher the more its recent motion looks to the discriminator like a demonstration. The demonstrations are thus never tracked step by step: the style reward draws exploration toward the demonstrated maneuvers, and the task reward decides which behavior is kept.

As \emph{demonstration resets}, intermediate states of the physics-replayed demonstrations (Sec.~\ref{sec:data}) initialize episodes in the middle of these maneuvers, for example with one carrier turned sideways or the load already pitched ahead of a corner, in the spirit of reference state initialization~\cite{peng2018deepmimic} and staged resets~\cite{xue2026doorman}. States along the paths of a planner are used in the same way (\emph{planner resets}). The planner is a sampling-based motion planner~\cite{kuffner2000rrtconnect} that uses the simulator's corridor geometry to find a collision-free path for the load, with each carrier kept in a default formation: at its end of the load, facing it. Both kinds of state contain the load and both robots, and they differ in how the robots are arranged around the load (Fig.~\ref{fig:exploration}, bottom): planner states have the robots in the default formation (dashed), whereas demonstration states have them in the formations that operators actually use at that point of the maneuver (solid). Either way, the teacher is given such states before it has ever reached them, unlike a buffer filled by its own rollouts.

\textbf{Stage 2: student distillation and co-training.} The visual student $\pi_S(a\mid o)$ is initialized from GR00T~N1.6~\cite{bjorck2025gr00t,nvidia2025gr00tn16}, a pretrained humanoid vision-language-action model. Both robots' RGB views are encoded as camera inputs; aligned LiDAR, proprioception, grasp configuration, and goal information enter through additional tokens. The action expert is adapted to predict the shared 27-D load-and-carrier action. Interactive imitation learning with DAgger~\cite{ross2011dagger} collects student-visited states in rendered simulation, with randomized appearance, lighting, and sensor calibration, and labels them with the privileged teacher's actions. The student is also trained directly on the physics-replayed VR demonstrations and on real two-robot teleoperation. The student is trained with the loss $\mathcal{L} = \mathcal{L}_{\mathrm{flow}} + \lambda_c\mathcal{L}_{\mathrm{clear}} + \lambda_\ell\mathcal{L}_{\mathrm{load}} + \lambda_v\mathcal{L}_{\mathrm{value}}$, which combines flow-matching imitation of action chunks with supervision of the clearance head, the load head, and the team value. Additional optimization details are given in the supplementary material.

\textbf{Stage 3: RL post-training.} The student observes less than the teacher, so the best behavior available to it can differ from what it imitates. For instance, the student may learn to slow down before a turn when the available observations provide uncertain clearance estimates. We therefore let the student improve on its own simulation rollouts with advantage-conditioned policy improvement~\cite{pi2025recap}. A single value function estimates the return of the joint team behavior; the policy is trained conditioned on whether each stored chunk had a positive advantage, and is deployed conditioned on a positive one. Demonstration replay maintains the coordinated behaviors learned from VR and teleoperation.

\begin{figure*}[t]
  \centering
    \includegraphics[width=\textwidth]{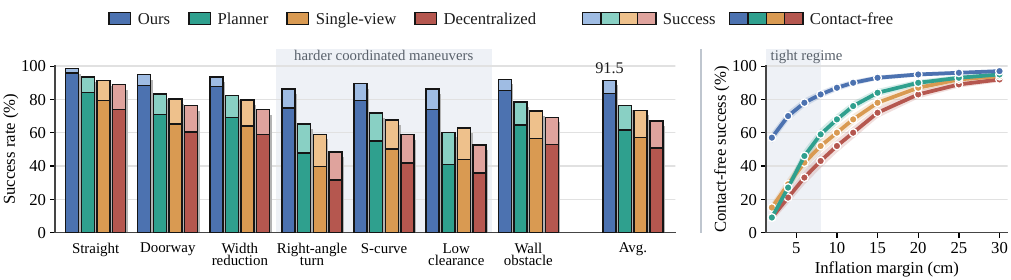}
  \caption{\textbf{Cooperative transport in simulation.}
Left: Overall success (light) and contact-free success (dark) across corridor types on unseen layouts and loads.
Right: Contact-free success versus inflation margin (mean $\pm$ std. over three seeds).}
  \label{fig:main}
\end{figure*}

\subsection{Corridors and Demonstrations}
\label{sec:data}

Training and evaluation cover seven corridor types: straight passage, doorway, width reduction, right-angle turn, S-curve, low clearance, and wall obstacle (Fig.~\ref{fig:data}). Each is generated at three difficulty levels (wide, medium, tight), which set the corridor width, the door width, and the headroom. Procedural variation changes passage dimensions, appearance, and load geometry, with held-out layouts and loads reserved for evaluation. Corridors are placed in 50 environments across 10 themes, and loads are 200 furniture models in 20 categories.
 Whether a load fits depends on its own shape as much as on the corridor, so we also measure the difficulty of a corridor--load pair by its \emph{inflation margin}: the largest uniform inflation of all obstacles for which the planner of Sec.~\ref{sec:learning} still finds a path.

In VR, two operators each control one virtual humanoid in a shared scene. Their paired demonstrations record how they reorient the load and reposition the carriers in confined spaces. These demonstrations supply the teacher's maneuver prior directly. Because the virtual robots are moved kinematically, each demonstration is also replayed in physics through $\pi_{\mathrm{wbc}}$; replays that complete provide the teacher's reset states and the student's imitation labels, together with rendered camera and LiDAR observations. For real teleoperation, two operators each command one physical robot using the same whole-body tracking interface, providing real sensor observations and execution data. Details of the VR interface, procedural generator, physics replay, and dataset composition are provided in the supplementary material.

\section{Experiments}
\label{sec:experiments}
 
We evaluate \method{} against three autonomous baselines in simulation and compare its real-world performance with two operators jointly controlling a pair of \robot{} humanoids. Ablations examine shared-load actions, cooperative perception, paired demonstrations, and student post-training.
 
\subsection{Experimental Setup}
\label{sec:exp_setup}
 
\noindent\textbf{Scenarios and protocol.} In simulation, we evaluate all seven corridor types at wide,
medium, and tight difficulty levels using held-out layouts
and loads. Trials begin 2\,m before the corridor entrance,
with the goal 1\,m beyond its exit. Hardware evaluation
focuses on the tight setting to test coordination under
restricted clearance, with 20 trials per method across
the tested corridor--load configurations.

\textbf{Robot system and transfer.} Two \robot{} humanoids each supply forward RGB, LiDAR, and proprioception. A shared policy aligns both robots' observations in the load frame and replans at \replanhz\,Hz; onboard whole-body controllers track hand and pelvis targets at \trackerhz\,Hz. We train in Isaac Lab with randomized appearance, lighting, and sensor calibration, as detailed in Sec.~\ref{sec:learning}. No external tracking is used for policy input.

\textbf{Metrics.} A trial succeeds when the load reaches the goal with both robots upright. \emph{Contact-free success} additionally requires no environment contact by either carrier or the load. For real-world trials, we also report mean completion time.

\textbf{Baselines.} In simulation, \emph{Decentralized} runs a local-observation student independently on each robot; \emph{Single-view} uses the centralized policy with only the leading robot's sensors; and \emph{Planner} uses RRT-Connect~\cite{kuffner2000rrtconnect} on a 5\,cm grid updated from both aligned LiDARs, checking the true load mesh's convex hull and carrier capsules. It replans online, expressing load and default-formation carrier targets through Eq.~\eqref{eq:action}. All share the same whole-body controller $\pi_{\mathrm{wbc}}$. On hardware, two expert operators each teleoperate one robot through that controller; this comparison is evaluated only in the real world.

\subsection{Cooperative Transport in Confined Corridors}
\label{sec:exp_main}
 
Fig.~\ref{fig:main} (left) compares simulation success and contact-free success across corridor types. \method{} achieves 91.5\% success and 83.6\% contact-free success overall, versus 76.4\% and 61.7\% for Planner, the strongest baseline. On right-angle turns and low-clearance passages, \method{} achieves 86.3\% and 86.2\% success, compared with 65.3\% and 60.2\% for Planner. These reorientations favor adaptive carrier
configurations over Planner's fixed formation,
despite its access to ground-truth load geometry.

The advantage grows as clearance decreases (Fig.~\ref{fig:main}, right). At an inflation margin of 30\,cm, the methods are within five percentage points; at 2\,cm, \method{} retains 57\% contact-free success, while every baseline is at or below 15\%. On tight-level corridors, \method{} achieves 69.6\% contact-free success, compared with 33.4\% for Planner, 30.4\% for Single-view, and 23.0\% for Decentralized. Tight clearances amplify mapping and tracking errors; the following ablations test the corresponding perception and coordination components.

\subsection{Real-World Deployment}
\label{sec:exp_real}

\begin{figure*}[t]
  \centering
  \begin{minipage}[c]{4.20in}
  
  \includegraphics[width=\linewidth]{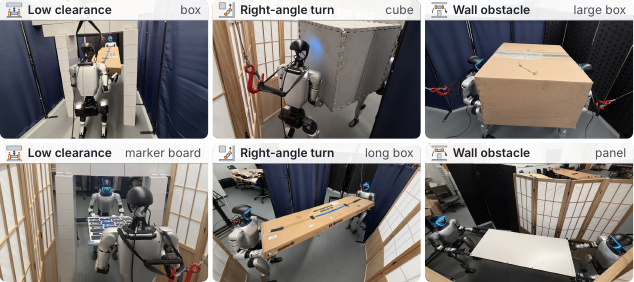}  
  \end{minipage}\hfill
  \begin{minipage}[c]{2.55in}
    \includegraphics[width=\linewidth]{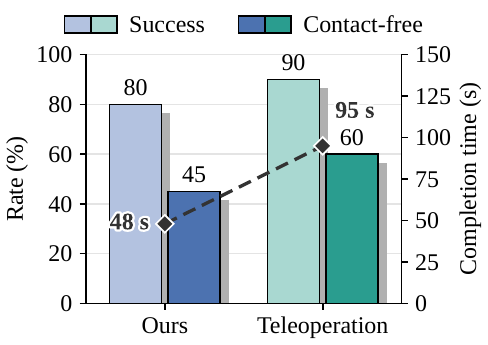}
  \end{minipage}
  \caption{\textbf{Real-world cooperative transport.} Left: \method{} deployed on two humanoids. Right: comparison with expert teleoperation.}
  \label{fig:real}
\end{figure*}

We deploy \method{} on two physical \robot{} humanoids using onboard sensing and the shared policy (Fig.~\ref{fig:real}). Across 20 trials per method, \method{} succeeds in 16 (80\%) and completes 9 without environment contact (45\%). Expert teleoperation succeeds in 18 (90\%), with 12 contact-free trials (60\%). Mean completion times are 48\,s for \method{} and 95\,s for teleoperation. Thus, \method{} achieves faster observed execution, while the human-operated pair has higher success and contact-free success on this evaluation.
 
\subsection{Effect of the Formulation and Load Perception}
\label{sec:exp_formulation}
 
\begin{figure}[!htbp]
  \centering
  \includegraphics[width=\linewidth]{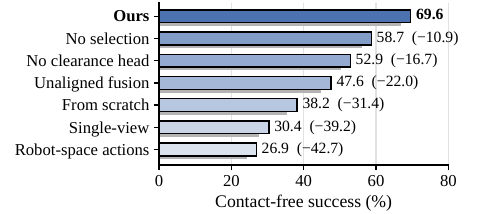}
  \caption{\textbf{Formulation ablation.} Contact-free success at the tight level; drop from the full method in parentheses.}
  \label{fig:abl_form}
\end{figure}
 
We evaluate the main design choices on tight corridors (Fig.~\ref{fig:abl_form}). \emph{Robot-space actions} replaces Eq.~\eqref{eq:action} with independent pelvis and hand targets for each robot. \emph{Unaligned fusion} encodes both robots' sensors as separate views without load-frame alignment. \emph{No selection} executes the first complete $H$-step chunk generated by the action expert, before clearance ranking; \emph{no clearance head} removes the learned clearance predictor, and \emph{from scratch} removes the pretrained student initialization.

Robot-space actions produce the largest decrease in contact-free success, from 69.6\% to 26.9\%, showing the importance of a consistent commanded load trajectory. Unaligned fusion reaches 47.6\%, while removing clearance-guided selection yields 58.7\%. The remaining ablations further support the benefits of clearance prediction and pretrained initialization.
 
\begin{figure}[!htbp]
  \centering
  \includegraphics[width=\linewidth]{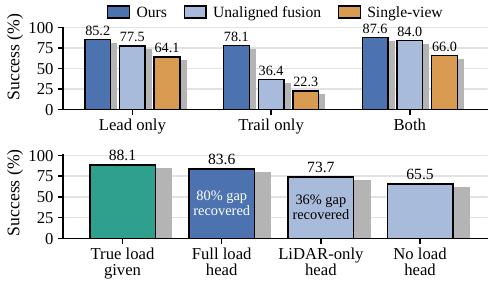}
  \caption{\textbf{Perception ablations.} Top: success by which robot sees the obstacle closest to the team. Bottom: success on held-out loads.}
  \label{fig:abl_view}
\end{figure}
 
\textbf{Complementary visibility.} We group episodes by which robot can see the obstacle closest to the team (Fig.~\ref{fig:abl_view}, top). The benefit of load-frame alignment is most pronounced when only the trailing robot observes it: our method succeeds in 78.1\% of episodes, compared with 36.4\% for unaligned fusion and 22.3\% for Single-view. When both robots see the obstacle, the two fusion methods are much closer.

\textbf{Load perception.} On held-out objects (Fig.~\ref{fig:abl_view}, bottom), RGB--LiDAR load perception achieves 83.6\% success, compared with 73.7\% for the LiDAR-only load head and 88.1\% with ground-truth load geometry. The multimodal head recovers about 80\% of the gap between having no load head and the ground-truth upper bound, supporting complementary appearance and range cues. RGB complements sparse LiDAR measurements by capturing
fine object boundaries and appearance cues for handling
constraints, improving load perception under partial
observability.
 
\subsection{Effect of Paired Demonstrations and Post-Training}
\label{sec:exp_data}

\begin{figure}[!t]
  \centering
  \includegraphics[width=\linewidth]{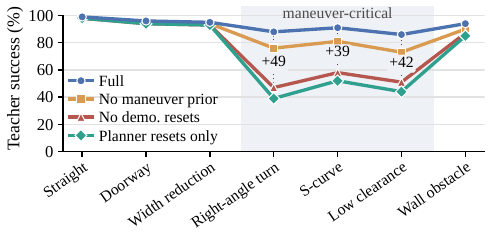}
  \caption{\textbf{Effect of paired demonstrations on the teacher.} Numbers give the gain of the full teacher over planner resets only.}
  \label{fig:abl_teacher}
\end{figure}

The style reward favors demonstrated maneuvers; demonstration resets initialize intermediate load--carrier states from physics replays, after which the teacher acts autonomously. We compare four teachers with the same task reward and PPO setup: the full method; \emph{planner resets only}, without VR data; \emph{no maneuver prior}, retaining demonstration resets but omitting the style reward; and \emph{no demonstration resets}, retaining the style reward (Fig.~\ref{fig:abl_teacher}). At the tight level, all variants are within two points on straight passages, doorways, and width reductions. On right-angle turns, S-curves, and low-clearance passages, the full teacher succeeds in 88\%, 91\%, and 86\%, compared with 39\%, 52\%, and 44\% using planner resets alone. Removing demonstration resets leads to larger losses than removing only the style reward.

We further count \emph{formation changes}, defined by a carrier heading change of more than 45$^\circ$ relative to the load. These occur in 64\% of full-teacher episodes and 12\% with planner resets only, compared with 71\% of VR demonstrations, consistent with demonstrations supplying useful intermediate formations.

\begin{figure}[!t]
  \centering
  \includegraphics[width=\linewidth]{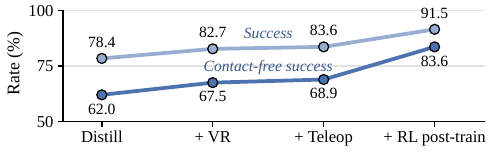}
  \caption{\textbf{Student after each training stage.}}
  \label{fig:abl_stages}
\end{figure}
 
Fig.~\ref{fig:abl_stages} traces the student across training stages. Distillation starts at 78.4\% success and 62.0\% contact-free success; VR replays and real teleoperation provide direct action-chunk supervision, improving both metrics. RL post-training yields the largest contact-free gain, reaching 91.5\% success and 83.6\% contact-free success.

\section{Related Work}
\label{sec:related}
 
\subsection{Cooperative Transport with Legged Robots}
 
Carrying a shared load with legged robots was first addressed with model-based control~\cite{wu2016symmetry,wen2024coordinated,hawley2019control}. Closest to our setting, Rioux~\etal~\cite{rioux2017cooperative} moved an object around obstacles with two small humanoids and head-mounted RGB-D sensors, using a pipeline of mapping, planning, and synchronization. Learning-based methods have since shown decentralized transport for bipeds~\cite{pandit2024decmbc}, quadrupeds~\cite{pandit2026decplm,cao2026paint}, and humanoid teams~\cite{pandit2026decmht}; these policies track a commanded object motion and do not perceive the scene. Pairs of quadrupeds pass narrow gates with safe RL using motion capture~\cite{lei2026dual}, and physics-based characters learn multi-agent carrying from privileged state~\cite{gao2024coohoi,lionar2026teamhoi}. A parallel line of work pairs a humanoid with a person~\cite{agravante2019human,bethala2025h2compact,du2025cola}, extended in~\cite{zhang2026c2c} to two humanoids and a human with partner state from motion capture. In contrast, \method{} learns one policy for two humanoids from onboard cameras and LiDAR, with no human in the loop and no external localization.
 
\subsection{Perceptive Control and Visual Sim-to-Real}
 
Perceptive controllers have taken single humanoids from parkour~\cite{zhuang2024parkour} to full 3-D constraints, using voxelized LiDAR~\cite{ben2026gallant}, potential fields around the body~\cite{xue2026humanoidpf}, or scene-aligned human motion~\cite{passage2026,wang2026mtc}. For manipulation, VIRAL~\cite{he2026viral} and DoorMan~\cite{xue2026doorman} distill privileged RL teachers into RGB students under heavy visual randomization. We adopt this recipe and extend it to a physically coupled pair with two views, where clearance is defined for the robot--load--robot system and exploration is seeded by human pairs.
 
\subsection{Human Data and RL Fine-Tuning}
 
Pretrained vision-language-action models~\cite{black2024pi0,bjorck2025gr00t,wei2026psi0} are adapted to humanoids with teleoperated data~\cite{ze2025twist,he2024omnih2o,ben2025homie,luo2025sonic} and, increasingly, egocentric human demonstrations~\cite{qiu2025humanoid,shi2026egohumanoid}. Our human data comes from pairs of operators who jointly steer two virtual humanoids coupled through the object they carry, and it serves both as a prior for RL and as imitation data. RL fine-tuning then refines a policy through interaction: by conditioning on advantages~\cite{pi2025recap,sala2026deed}, by learning in a compact latent space~\cite{gu2026haf}, by updating planner and controller jointly~\cite{gu2026refinedp}, or by generating better data~\cite{xiao2025pld}. We apply advantage conditioning to a two-robot policy with a single team value function.
\section{Conclusion}
\label{sec:conclusion}

We introduced \method{}, a vision-based framework for
cooperative humanoid transport through confined spaces.
By aligning complementary observations in a shared load
frame and jointly predicting load and carrier motions,
the policy supports coordinated maneuvers under partial
visibility. Paired VR demonstrations provide maneuver
priors, intermediate states for teacher training, and
direct supervision for the student. Experiments across
seven simulated corridor types show the largest benefits
in tight turns and low-clearance passages. Physical
experiments demonstrate autonomous transport with shorter
observed completion times than expert teleoperation,
while teleoperation achieves higher success rates.
These findings motivate extending shared-load coordination
to dynamic environments and changing grasp configurations.

{
    \small
    \bibliographystyle{ieeenat_fullname}
    \bibliography{main}

@String(CVPR= {IEEE Conf. Comput. Vis. Pattern Recog.})

@String(CVPR  = {CVPR})

@article{luo2025sonic,
  author = {Luo, Zhengyi and Yuan, Ye and Wang, Tingwu and Li, Chenran and Casta{\~n}eda, Fernando and Chen, Sirui and Cao, Zi-Ang and Li, Jiefeng and Minor, David and Ben, Qingwei and Park, Jinhyung and Sami, David and Wang, Zi and Da, Xingye and Ding, Runyu and Hogg, Cyrus and Song, Lina and Lim, Edy and Jeong, Eugene and He, Tairan and Xue, Haoru and Xiao, Wenli and Yuen, Simon and Kautz, Jan and Chang, Yan and Iqbal, Umar and Fan, Linxi and Zhu, Yuke},
  title = {{SONIC}: Supersizing Motion Tracking for Natural Humanoid Whole-Body Control},
  journal = {Science Robotics},
  volume = {11},
  number = {117},
  pages = {eaed4592},
  year = {2026},
  doi = {10.1126/scirobotics.aed4592}
}

@inproceedings{du2025cola,
  title     = {Learning Human-Humanoid Coordination for Collaborative Object Carrying},
  author    = {Du, Yushi and Li, Yixuan and Jia, Baoxiong and Lin, Yutang and Zhou, Pei and Liang, Wei and Yang, Yanchao and Huang, Siyuan},
  booktitle = {IEEE International Conference on Robotics and Automation (ICRA)},
  year      = {2026},
  url       = {https://arxiv.org/abs/2510.14293}
}

@article{pandit2026decmht,
  author = {Pandit, Bikram and Gadde, Mohitvishnu S. and Shrestha, Aayam Kumar and Fern, Alan},
  title = {Learning Multi-Humanoid Pickup and Transport via Decentralized Object-Centric Control},
  journal = {arXiv preprint arXiv:2609.17824},
  year = {2026},
  url = {https://arxiv.org/abs/2609.17824}
}

@inproceedings{gao2024coohoi,
  title     = {{CooHOI}: Learning Cooperative Human-Object Interaction with Manipulated Object Dynamics},
  author    = {Gao, Jiawei and Wang, Ziqin and Xiao, Zeqi and Wang, Jingbo and Wang, Tai and Cao, Jinkun and Hu, Xiaolin and Liu, Si and Dai, Jifeng and Pang, Jiangmiao},
  booktitle = {Advances in Neural Information Processing Systems (NeurIPS)},
  year      = {2024}
}

@inproceedings{lionar2026teamhoi,
  title     = {{TeamHOI}: Learning a Unified Policy for Cooperative Human-Object Interactions with Any Team Size},
  author    = {Lionar, Stefan and Lee, Gim Hee},
  booktitle = {Proceedings of the IEEE/CVF Conference on Computer Vision and Pattern Recognition (CVPR)},
  year      = {2026}
}

@inproceedings{pandit2024decmbc,
  title     = {Learning Decentralized Multi-Biped Control for Payload Transport},
  author    = {Pandit, B. and Gupta, A. and Gadde, M. S. and Johnson, A. and Duan, H. and Dao, J. and Shrestha, A. K. and Fern, A.},
  booktitle = {Conference on Robot Learning (CoRL)},
  year      = {2024}
}

@inproceedings{pandit2026decplm,
  title     = {Multi-Quadruped Cooperative Object Transport: Learning Decentralized Pinch-Lift-Move},
  author    = {Pandit, Bikram and Shrestha, Aayam Kumar and Fern, Alan},
  booktitle = {IEEE International Conference on Robotics and Automation (ICRA)},
  year      = {2026}
}

@article{zhang2026c2c,
  title   = {Physical Contracts for Plug-Compatible Human-Humanoid Collaboration},
  author  = {Zhang, Hao and Li, Yisen and Geng, Ruize and Tseng, Yves and Niu, Yaru and Zhao, Ding and Tseng, H. Eric},
  journal = {arXiv preprint arXiv:2603.03768},
  year    = {2026}
}

@article{agravante2019human,
  title   = {Human-Humanoid Collaborative Carrying},
  author  = {Agravante, Don Joven and Cherubini, Andrea and Sherikov, Alexander and Wieber, Pierre-Brice and Kheddar, Abderrahmane},
  journal = {IEEE Transactions on Robotics},
  volume  = {35},
  number  = {4},
  pages   = {833--846},
  year    = {2019}
}

@inproceedings{bethala2025h2compact,
  title     = {{H$^2$-COMPACT}: Human-Humanoid Co-Manipulation via Adaptive Contact Trajectory Policies},
  author    = {Bethala, G. C. R. and Huang, H. and Pudasaini, N. and Ali, A. M. and Yuan, S. and Wen, C. and Tzes, A. and Fang, Y.},
  booktitle = {IEEE-RAS International Conference on Humanoid Robots (Humanoids)},
  pages     = {1004--1011},
  year      = {2025}
}

@article{cao2026paint,
  title = {{PAINT}: Partner-Agnostic Intent-Aware Cooperative Transport with Legged Robots},
  author = {Cao, Zhihao and An, Tianxu and Li, Chenhao and Coros, Stelian and Hutter, Marco},
  journal = {IEEE Robotics and Automation Letters},
  volume = {11},
  number = {11},
  pages = {12312--12319},
  year = {2026},
  doi = {10.1109/LRA.2026.3730304}
}

@article{hawley2019control,
  title   = {Control Framework for Cooperative Object Transportation by Two Humanoid Robots},
  author  = {Hawley, L. and Suleiman, W.},
  journal = {Robotics and Autonomous Systems},
  volume  = {115},
  pages   = {1--16},
  year    = {2019}
}

@article{rioux2017cooperative,
  title   = {Cooperative Vision-Based Object Transportation by Two Humanoid Robots in a Cluttered Environment},
  author  = {Rioux, Antoine and Esteves, Claudia and Hayet, Jean-Bernard and Suleiman, Wael},
  journal = {International Journal of Humanoid Robotics},
  volume  = {14},
  number  = {3},
  pages   = {1750018},
  year    = {2017}
}

@article{wu2016symmetry,
  title   = {Symmetry Position/Force Hybrid Control for Cooperative Object Transportation Using Multiple Humanoid Robots},
  author  = {Wu, M.-H. and Ogawa, S. and Konno, A.},
  journal = {Advanced Robotics},
  volume  = {30},
  number  = {2},
  pages   = {131--149},
  year    = {2016}
}

@article{wen2024coordinated,
  title   = {Coordinated Transport by Dual Humanoid Robots Using Distributed Model Predictive Control},
  author  = {Wen, S. and Shi, Z. and Li, H.},
  journal = {Biomimetics},
  volume  = {9},
  number  = {6},
  pages   = {332},
  year    = {2024}
}

@article{lei2026dual,
  title   = {Dual-Quadruped Collaborative Transportation in Narrow Environments via Safe Reinforcement Learning},
  author  = {Lei, Zhezhi and Bi, Zhihai and Wang, Wenxin and Ma, Jun},
  journal = {arXiv preprint arXiv:2602.16353},
  year    = {2026}
}

@article{xue2026humanoidpf,
  title   = {Collision-Free Humanoid Traversal in Cluttered Indoor Scenes},
  author  = {Xue, Han and Liang, Sikai and Zhang, Zhikai and Zeng, Zicheng and Liu, Yun and Lian, Yunrui and Wang, Jilong and Liu, Qingtao and Shi, Xuesong and Yi, Li},
  journal = {IEEE Robotics and Automation Letters},
  volume  = {11},
  number  = {8},
  pages   = {9183--9190},
  year    = {2026},
  doi     = {10.1109/LRA.2026.3703590}
}

@article{passage2026,
  title   = {{PASSAGE}: Scaling Scene-Aligned Motion Learning for Perceptive Humanoid Traversal in Cluttered Environments},
  author  = {Yuxuan Ma and Zicheng Zeng and Chunlin Peng and Zhoujian Li and Zetong Zhao and Zhikai Zhang and Yunrui Lian and Han Xue and Sikai Liang and Weiyi Zhu and Mulin Chen and Chenghuai Lin and Jiayu Zeng and Yanwei An and Songan Zhang and Jiayuan Gu and Jilong Wang and Jingbo Wang and He Wang and Li Yi},
  journal = {arXiv preprint arXiv:2609.18732},
  year    = {2026},
  doi     = {10.48550/arXiv.2609.18732},
  url     = {https://arxiv.org/abs/2609.18732}
}

@article{wang2026mtc,
  title   = {Moving Through Clutter: Scaling Data Collection and Benchmarking for {3D} Scene-Aware Humanoid Locomotion via Virtual Reality},
  author  = {Wang, Beichen and Lu, Yuanjie and Wang, Linji and Yu, Liuchuan and Xiao, Xuesu},
  journal = {arXiv preprint arXiv:2603.05993},
  year    = {2026}
}

@inproceedings{ben2026gallant,
  title     = {Gallant: Voxel Grid-based Humanoid Locomotion and Local-navigation across {3-D} Constrained Terrains},
  author    = {Ben, Qingwei and Xu, Botian and Li, Kailin and Jia, Feiyu and Zhang, Wentao and Wang, Jingping and Wang, Jingbo and Lin, Dahua and Pang, Jiangmiao},
  booktitle = {Proceedings of the IEEE/CVF Conference on Computer Vision and Pattern Recognition (CVPR)},
  pages     = {28086--28095},
  year      = {2026}
}

@inproceedings{zhuang2024parkour,
  title     = {Humanoid Parkour Learning},
  author    = {Zhuang, Ziwen and Yao, Shenzhe and Zhao, Hang},
  booktitle = {Conference on Robot Learning (CoRL)},
  year      = {2024}
}

@inproceedings{he2026viral,
  title     = {{VIRAL}: Visual Sim-to-Real at Scale for Humanoid Loco-Manipulation},
  author    = {He, Tairan and Wang, Zi and Xue, Haoru and Ben, Qingwei and Luo, Zhengyi and Xiao, Wenli and Yuan, Ye and Da, Xingye and Casta{\~n}eda, Fernando and Sastry, Shankar and Liu, Changliu and Shi, Guanya and Fan, Linxi and Zhu, Yuke},
  booktitle = {Proceedings of the IEEE/CVF Conference on Computer Vision and Pattern Recognition (CVPR)},
  year      = {2026}
}

@inproceedings{xue2026doorman,
  title     = {Opening the Sim-to-Real Door for Humanoid Pixel-to-Action Policy Transfer},
  author    = {Xue, Haoru and He, Tairan and Wang, Zi and Ben, Qingwei and Xiao, Wenli and Luo, Zhengyi and Da, Xingye and Casta{\~n}eda, Fernando and Shi, Guanya and Sastry, Shankar and Fan, Linxi and Zhu, Yuke},
  booktitle = {Proceedings of the IEEE/CVF Conference on Computer Vision and Pattern Recognition (CVPR)},
  pages     = {6642--6652},
  year      = {2026}
}

@inproceedings{ze2025twist,
  title     = {{TWIST}: Teleoperated Whole-Body Imitation System},
  author    = {Ze, Yanjie and Chen, Zixuan and Ara{\'u}jo, Jo{\~a}o Pedro and Cao, Zi-ang and Peng, Xue Bin and Wu, Jiajun and Liu, C. Karen},
  booktitle = {Conference on Robot Learning (CoRL)},
  year      = {2025}
}

@inproceedings{he2024omnih2o,
  title     = {{OmniH2O}: Universal and Dexterous Human-to-Humanoid Whole-Body Teleoperation and Learning},
  author    = {He, Tairan and Luo, Zhengyi and He, Xialin and Xiao, Wenli and Zhang, Chong and Zhang, Weinan and Kitani, Kris and Liu, Changliu and Shi, Guanya},
  booktitle = {Conference on Robot Learning (CoRL)},
  year      = {2024}
}

@inproceedings{ben2025homie,
  title     = {{HOMIE}: Humanoid Loco-Manipulation with Isomorphic Exoskeleton Cockpit},
  author    = {Ben, Qingwei and Jia, Feiyu and Zeng, Jia and Dong, Junting and Lin, Dahua and Pang, Jiangmiao},
  booktitle = {Robotics: Science and Systems (RSS)},
  year      = {2025}
}

@article{bjorck2025gr00t,
  title   = {{GR00T N1}: An Open Foundation Model for Generalist Humanoid Robots},
  author  = {Bjorck, Johan and Casta{\~n}eda, Fernando and Cherniadev, Nikita and Da, Xingye and Ding, Runyu and Fan, Linxi and others},
  journal = {arXiv preprint arXiv:2503.14734},
  year    = {2025}
}

@inproceedings{wei2026psi0,
  title     = {{$\Psi_0$}: An Open Foundation Model Towards Universal Humanoid Loco-Manipulation},
  author    = {Wei, Songlin and Jing, Hongyi and Li, Boqian and Zhao, Zhenyu and Mao, Jiageng and Ni, Zhenhao and He, Sicheng and others},
  booktitle = {Robotics: Science and Systems (RSS)},
  year      = {2026}
}

@inproceedings{black2024pi0,
  title = {{$\pi_0$}: A Vision-Language-Action Flow Model for General Robot Control},
  author = {Black, Kevin and Brown, Noah and Driess, Danny and Esmail, Adnan and Equi, Michael and Finn, Chelsea and Fusai, Niccolo and Groom, Lachy and Hausman, Karol and Ichter, Brian and Jakubczak, Szymon and Jones, Tim and Ke, Liyiming and Levine, Sergey and Li-Bell, Adrian and Mothukuri, Mohith and Nair, Suraj and Pertsch, Karl and Shi, Lucy Xiaoyang and Tanner, James and Vuong, Quan and Walling, Anna and Wang, Haohuan and Zhilinsky, Ury},
  booktitle = {Robotics: Science and Systems (RSS)},
  year = {2025},
  url = {https://www.roboticsproceedings.org/rss21/p010.html}
}

@inproceedings{qiu2025humanoid,
  title = {Humanoid Policy $\sim$ Human Policy},
  author = {Qiu, Ri-Zhao and Yang, Shiqi and Cheng, Xuxin and Chawla, Chaitanya and Li, Jialong and He, Tairan and Yan, Ge and Yoon, David J. and Hoque, Ryan and Paulsen, Lars and Yang, Ge and Zhang, Jian and Yi, Sha and Shi, Guanya and Wang, Xiaolong},
  booktitle = {Proceedings of the 9th Conference on Robot Learning},
  series = {Proceedings of Machine Learning Research},
  volume = {305},
  pages = {2888--2906},
  publisher = {PMLR},
  year = {2025},
  url = {https://proceedings.mlr.press/v305/qiu25a.html}
}

@inproceedings{shi2026egohumanoid,
  title     = {{EgoHumanoid}: Unlocking In-the-Wild Loco-Manipulation with Robot-Free Egocentric Demonstration},
  author    = {Shi, Modi and Peng, Shijia and Chen, Jin and Jiang, Haoran and Li, Tianyu and Luo, Ping and Huang, Di and Li, Hongyang and Chen, Li},
  booktitle = {Robotics: Science and Systems (RSS)},
  year      = {2026}
}

@inproceedings{ross2011dagger,
  title     = {A Reduction of Imitation Learning and Structured Prediction to No-Regret Online Learning},
  author    = {Ross, St{\'e}phane and Gordon, Geoffrey and Bagnell, Drew},
  booktitle = {Proceedings of the Fourteenth International Conference on Artificial Intelligence and Statistics (AISTATS)},
  pages     = {627--635},
  year      = {2011}
}

@article{schulman2017ppo,
  title   = {Proximal Policy Optimization Algorithms},
  author  = {Schulman, John and Wolski, Filip and Dhariwal, Prafulla and Radford, Alec and Klimov, Oleg},
  journal = {arXiv preprint arXiv:1707.06347},
  year    = {2017}
}

@article{peng2021amp,
  title   = {{AMP}: Adversarial Motion Priors for Stylized Physics-Based Character Control},
  author  = {Peng, Xue Bin and Ma, Ze and Abbeel, Pieter and Levine, Sergey and Kanazawa, Angjoo},
  journal = {ACM Transactions on Graphics},
  volume  = {40},
  number  = {4},
  year    = {2021}
}

@article{peng2018deepmimic,
  title   = {{DeepMimic}: Example-Guided Deep Reinforcement Learning of Physics-Based Character Skills},
  author  = {Peng, Xue Bin and Abbeel, Pieter and Levine, Sergey and van de Panne, Michiel},
  journal = {ACM Transactions on Graphics},
  volume  = {37},
  number  = {4},
  year    = {2018}
}

@article{mittal2025isaaclab,
  title   = {{Isaac Lab}: A {GPU}-Accelerated Simulation Framework for Multi-Modal Robot Learning},
  author  = {Mittal, Mayank and Roth, Pascal and Tigue, James and Richard, Antoine and Zhang, Octi and others},
  journal = {arXiv preprint arXiv:2511.04831},
  year    = {2025}
}

@inproceedings{zhou2019continuity,
  title     = {On the Continuity of Rotation Representations in Neural Networks},
  author    = {Zhou, Yi and Barnes, Connelly and Lu, Jingwan and Yang, Jimei and Li, Hao},
  booktitle = {Proceedings of the IEEE/CVF Conference on Computer Vision and Pattern Recognition (CVPR)},
  pages     = {5745--5753},
  year      = {2019}
}

@article{pi2025recap,
  title   = {{$\pi^{*}_{0.6}$}: a {VLA} That Learns From Experience},
  author  = {{Physical Intelligence} and others},
  journal = {arXiv preprint arXiv:2511.14759},
  year    = {2025}
}

@inproceedings{sala2026deed,
  title = {Closing the Lab-to-Store Gap: A Data-Efficient Post-Training and Experience-Driven Learning {VLA} Framework for Retail Humanoids},
  author = {Sala Sis{\'o}, Roger and Silv{\'e}rio, Tiago and Sand, Jakob and Le, Tran Nguyen},
  booktitle = {2026 IEEE-RAS 25th International Conference on Humanoid Robots (Humanoids)},
  publisher = {IEEE},
  year = {2026},
  note = {Accepted/In press}
}

@article{gu2026haf,
  title   = {{HAF}: Adapting Generalist {VLAs} to Humanoid Whole-Body Loco-manipulation via Hierarchical Action Flow and Spectral Latent {RL}},
  author  = {Gu, Langzhe and Hou, Chengkai and Li, Meng and Wang, Xinhua and Liu, Jiaming and Lv, Xinyuan and Zhang, Bowei and Bai, Shuanghao and Li, Guangrun and He, Jingyang and Dai, Gaole and Ding, Ziluo and Xu, Zhiyuan and Cheng, Kuan and Tang, Jian and Che, Zhengping and Zhang, Shanghang},
  journal = {arXiv preprint arXiv:2608.16837},
  year    = {2026}
}

@article{gu2026refinedp,
  title = {{REFINE-DP}: Diffusion Policy Fine-Tuning for Humanoid Loco-Manipulation via Reinforcement Learning},
  author = {Gu, Zhaoyuan and Chen, Yipu and Chai, Zimeng and Cueva, Alfred and Nguyen, Thong and Wu, Yifan and Xue, Huishu and Kim, Minji and Legene, Isaac and Liu, Fukang and Kim, KyoungMok and Barula, Ayan and Chen, Yongxin and Zhao, Ye},
  journal = {IEEE Robotics and Automation Letters},
  volume = {11},
  number = {10},
  pages = {11617--11624},
  year = {2026},
  doi = {10.1109/LRA.2026.3723742}
}

@inproceedings{xiao2025pld,
  title     = {Self-Improving Vision-Language-Action Models with Data Generation via Residual {RL}},
  author    = {Xiao, Wenli and Lin, Haotian and Peng, Andy and Xue, Haoru and He, Tairan and Xie, Yuqi and Hu, Fengyuan and Wu, Jimmy and Luo, Zhengyi and Fan, Linxi and Shi, Guanya and Zhu, Yuke},
  booktitle = {International Conference on Learning Representations},
  year      = {2026}
}

@misc{nvidia2025gr00tn16,
  title        = {{GR00T N1.6}: An Improved Open Foundation Model for Generalist Humanoid Robots},
  author       = {{NVIDIA GEAR Team}},
  year         = {2025},
  howpublished = {\url{https://research.nvidia.com/labs/gear/gr00t-n1_6/}},
  note         = {Released December 2025}
}

@inproceedings{kuffner2000rrtconnect,
  title     = {{RRT-Connect}: An Efficient Approach to Single-Query Path Planning},
  author    = {Kuffner, James J. and LaValle, Steven M.},
  booktitle = {IEEE International Conference on Robotics and Automation (ICRA)},
  pages     = {995--1001},
  year      = {2000}
}
}


\end{document}